\documentclass[runningheads]{llncs}
\usepackage{graphicx}
\usepackage{booktabs,longtable,array}
\usepackage[table]{xcolor}
\usepackage{fvextra} 
\usepackage{tcolorbox} 
\tcbset{
    colback=gray!5,
    colframe=black,
    boxrule=0.8pt,
    arc=2pt,
    left=4pt,
    right=4pt,
    top=4pt,
    bottom=4pt
}
\usepackage{amsmath}
\definecolor{baselinegray}{gray}{0.95}
\newcommand{\CatSmall}{$\leq$8B}
\newcommand{\CatMid}{8--35B}
\newcommand{\CatLarge}{$>$35B}

\begin{document}
\title{ICDAR2026 Competition on Multimodal Reasoning over Documents in Multiple Domains}
\titlerunning{ICDAR2026 DocVQA Competition}

\author{Artemis Llabrés\orcidID{0000-0002-6128-1796} \and
Marc Serra Ortega\orcidID{0009-0005-3427-184X} \and
Tomàs Ockier\orcidID{0009-0002-0682-7589} \and
Samuel Ortega Cuadra\orcidID{0009-0006-4040-5004} \and
Amritpal Singh\orcidID{0009-0006-1960-2548} \and
Christos Georgakilas\orcidID{0009-0000-1461-1281} \and
Andrey Barsky\orcidID{0000-0002-6993-5969} \and
Ernest Valveny\orcidID{0000-0002-0368-9697} \and
Dimosthenis Karatzas\orcidID{0000-0001-8762-4454}}

\authorrunning{A. Llabrés et al.}

\institute{Computer Vision Center and Universitat Autònoma de Barcelona \\
\email{\{allabres, mserrao, tockier, sortegac\}@cvc.uab.cat}}
\maketitle              
\begin{abstract}
In this report we present results of the ICDAR2026 Competition on Multimodal Reasoning over Documents in Multiple Domains. This competition aimed to advance research in document understanding through the task of Visual Question Answering (VQA). Building upon previous DocVQA benchmarks, this competition introduces challenging reasoning questions over a diverse collection of documents spanning eight domains, including business reports, scientific papers, slides, posters, maps, comics, infographics, and engineering drawings. The competition concluded with 20 valid submissions from 8 teams spanning zero-shot VLMs, OCR and parser-augmented pipelines, agentic retrieval systems, multi-agent ensembles, and fine-tuned multimodal models. The results show that the strongest systems move beyond single-pass prompting and instead rely on structured evidence extraction, retrieval, verification, and orchestration across multiple components.
\end{abstract}

\section{Introduction}
\begin{figure}[ht]
    \centering
    \includegraphics[width=0.8\linewidth]{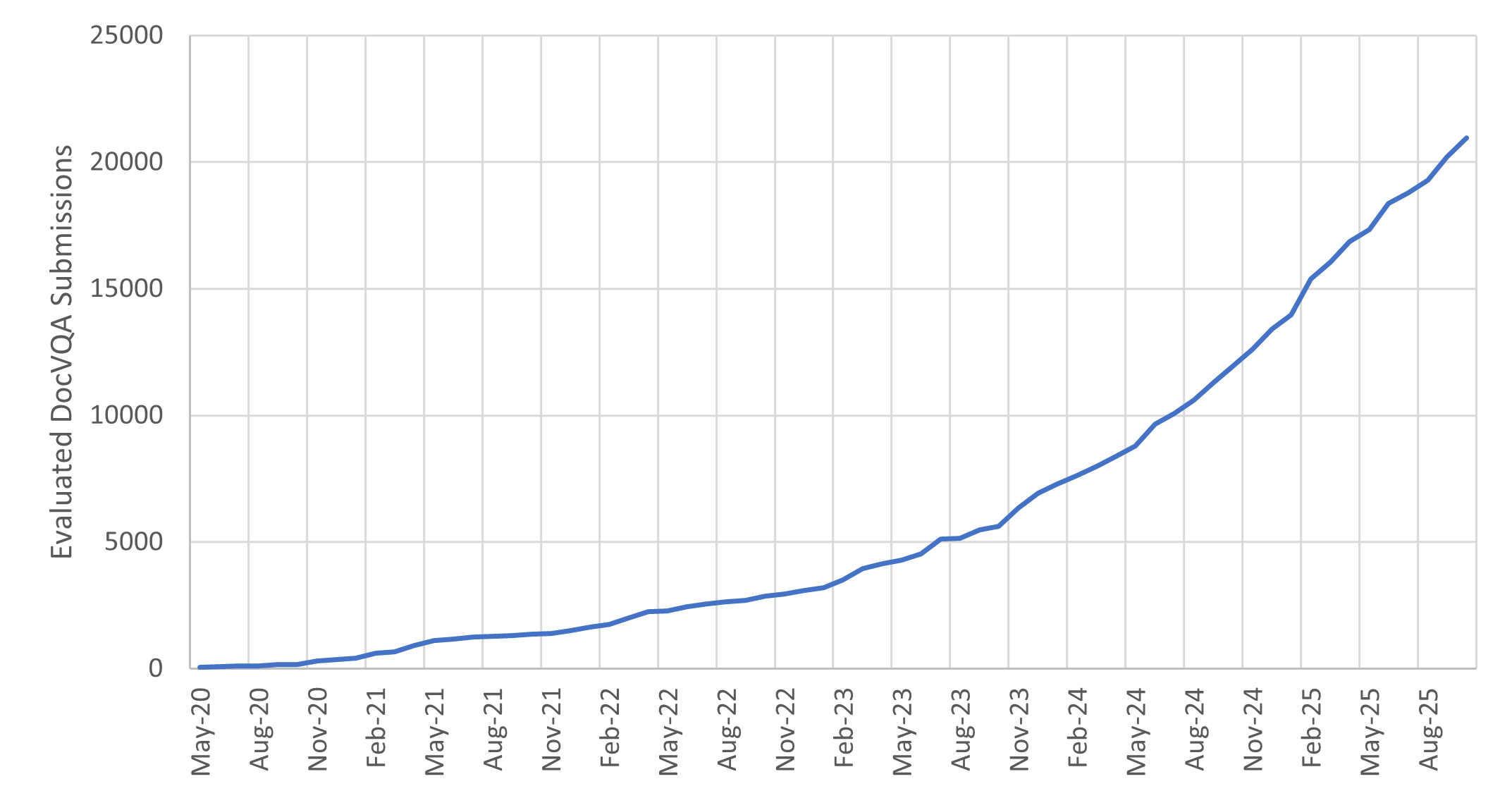}
    \caption{Evolution of submitted methods to the DocVQA competitions.}
    \label{fig:rrc}
\end{figure}

The task of Document VQA, introduced by members of our team in 2020~\cite{mathew2021docvqa}, marked a departure from classic information extraction tasks and a move towards end-to-end multimodal models for document understanding. The original benchmark focused on single page documents and on ``extractive'' questions. As such, the reasoning capabilities required from the first generation of DocVQA models were limited. The international community has nevertheless embraced this research direction, and to date we have evaluated more than $20,000$ submitted methods\footnote{Robust Reading Competition portal (\url{https://rrc.cvc.uab.es/})}, see Figure~\ref{fig:rrc}. In addition, the DocVQA benchmark has been one of the standard evaluation benchmarks used by multimodal models, such as OpenAI’s GPT4\footnote{Evaluation at: \url{https://openai.com/research/gpt-4}}, and Google Deepmind’s Gemini\footnote{Evaluation at: \url{https://deepmind.google/technologies/gemini}}.

Since the introduction of the \textit{DocVQA} dataset~\cite{mathew2021docvqa}, a large number of document–centric VQA benchmarks have been proposed, including \textit{InfographicsVQA}~\cite{mathew2022infographicvqa}, \textit{ChartQA}~\cite{masry2022chartqa}, \textit{SlideVQA}~\cite{tanaka2023slidevqa}, multi–page extensions such as \textit{MP\mbox{-}DocVQA}~\cite{tito2023hierarchical}, and broader document benchmarks such as \textit{MMLongBench\mbox{-}Doc}~\cite{ma2024mmlongbench}. While these resources were instrumental in establishing document understanding as a core multimodal research direction, many have reached performance saturation or focus on narrow problem formulations. For instance, the original \textit{DocVQA} and \textit{InfographicsVQA} benchmarks largely emphasize extractive questions over visually localized text, with current state-of-the-art systems achieving near-saturation performance (reported at approximately $98\%$ on DocVQA and $91\%$ on InfographicsVQA), therefore limiting their ability to measure continued progress. Similarly, \textit{ChartQA} restricts reasoning to chart images only, discarding broader page context, while \textit{SlideVQA} focuses solely on slide documents. Performance on the multi-page \textit{MP\mbox{-}DocVQA} benchmark has also reached high levels (around $88\%$) and primarily requires evidence from a single page, limiting the need for genuine long-range multimodal reasoning. Meanwhile, \textit{MMLongBench\mbox{-}Doc} targets a limited set of specialized domains (e.g., scientific reports, legal documents, and technical manuals), reducing coverage across diverse document types.

Beyond traditional documents, other multimodal domains remain underexplored in VQA settings. For comics, \textit{ComicsPAP}~\cite{vivoli2025comicspap} targets temporal and spatial reasoning through a Pick-a-Panel formulation rather than open-ended question answering. \textit{MangaVQA}~\cite{baek2026mangavqa} introduces VQA over manga content. In the geospatial domain, \textit{MapBench}~\cite{xing2025can} and \textit{MapTrace}~\cite{panagopoulou2025maptrace} provide valuable resources for map navigation and path tracing, yet they focus on instruction following and spatial trajectory reasoning rather than document-style question answering. These gaps highlight the need for a unified benchmark that spans multiple document genres and requires deeper multimodal reasoning beyond extraction, classification, or narrow domain specialization.

\begin{figure}[ht]
    \centering
    \includegraphics[width=\linewidth]{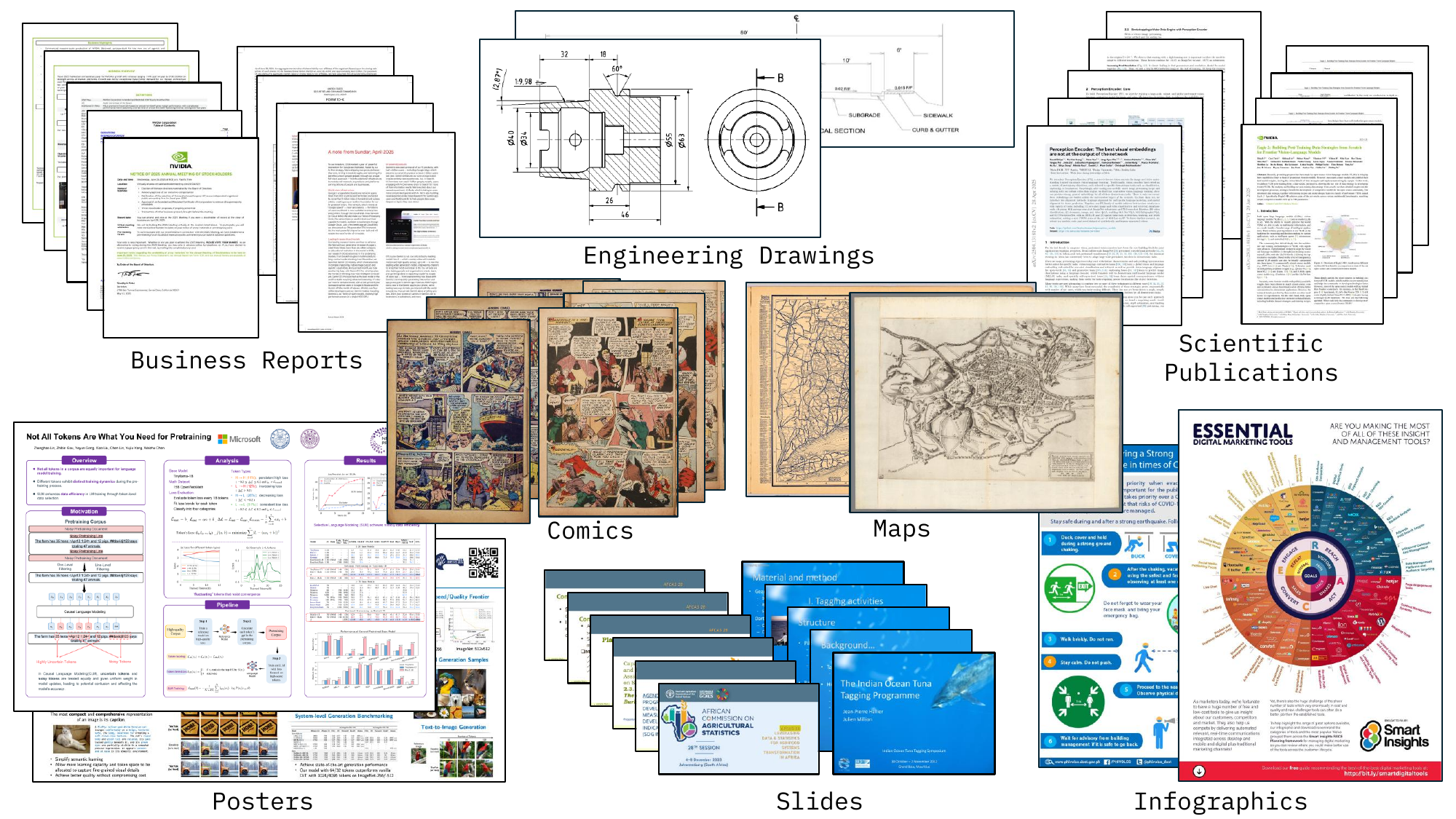}
    \caption{Overview of the eight document domains of the competition: business reports, scientific publications, slides, posters, maps, comics, infographics, and engineering drawings.}
    \label{fig:overview}
\end{figure}
The ICDAR2026 Competition on Multimodal Reasoning over Documents in Multiple Domains builds on the successful DocVQA series of competitions, and aims to evaluate models on documents from 8 different domains (see Figure \ref{fig:overview}), testing their multimodal reasoning abilities by introducing richer question types. This competition utilizes the DocVQA2026 dataset\footnote{Available at: \url{https://huggingface.co/datasets/VLR-CVC/DocVQA-2026}}. DocVQA2026 questions go beyond simple extraction, requiring reasoning across multiple sources of evidence along the document. The multimodal reasoning abilities that we expect evaluated methods to showcase include: spatial understanding (e.g. tested on maps, engineering drawings, and general layout tasks), temporal understanding (principally tested on comic stories), and multi-hop answers that require obtaining multiple evidences throughout the document (e.g. combining information from the text, tables, and figures). Therefore, the evaluated task is to answer questions posed over documents, with the model's final answer having to match one of the answer variations annotated in the ground truth.\newline

The competition was conducted through the Robust Reading Competition (RRC) platform with 20 valid submissions from 8 teams. The resulting leaderboard reflects a wide range of strategies, including direct zero-shot VLM prompting, OCR- and parser-augmented pipelines, retrieval-based and agentic systems, multi-agent ensembles, and fine-tuned multimodal models. As such, the competition serves both as an evaluation benchmark and as a snapshot of the current design space for multimodal reasoning over long, heterogeneous documents. The following sections present the dataset, the evaluation protocol, the submitted methods, and an analysis of the main findings of the competition.

\section{DocVQA2026 Dataset}

\begin{figure}
    \centering
    \includegraphics[width=\linewidth]{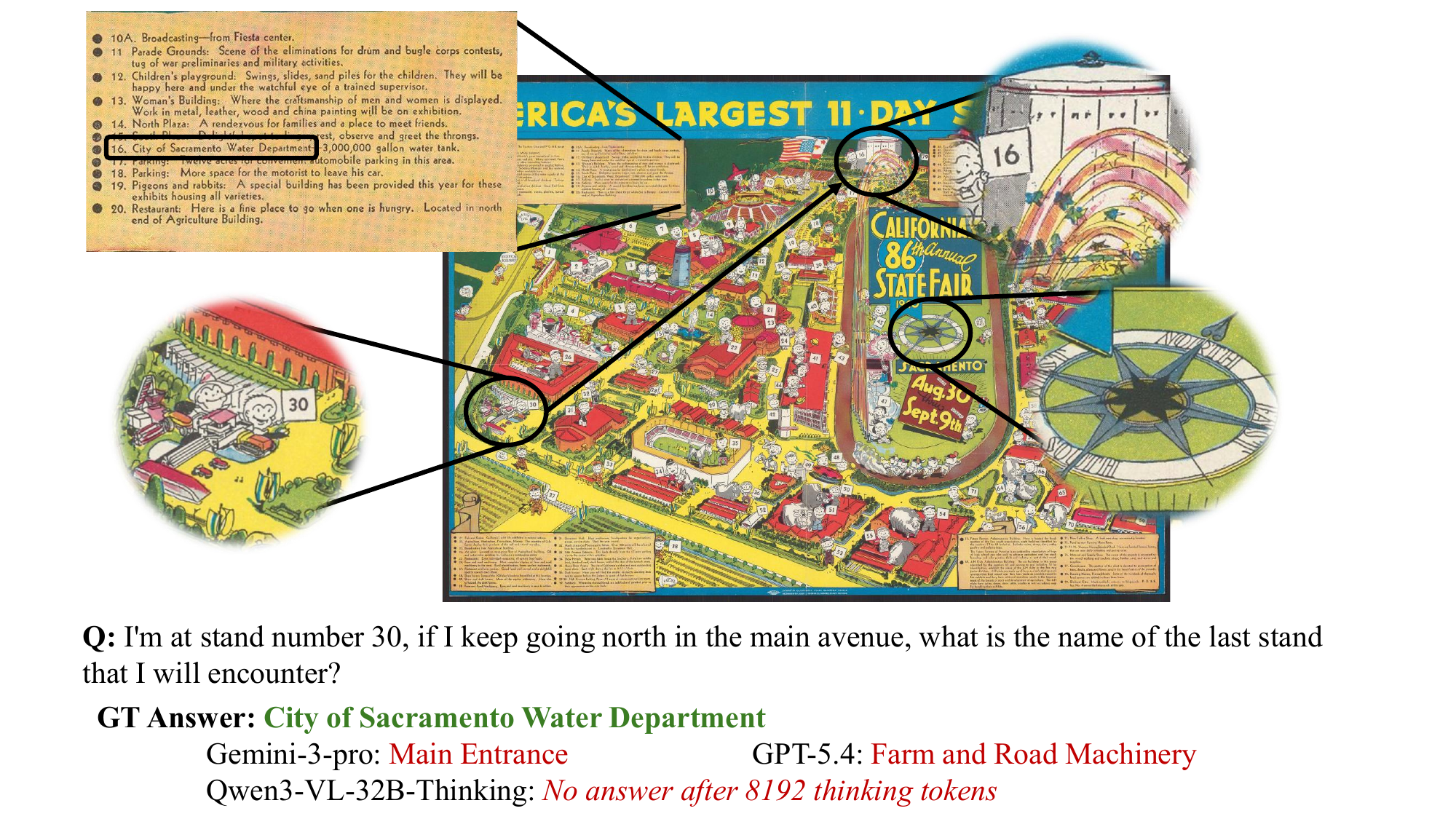}
    \caption{Sample question from the maps domain: to answer it, the model must locate stand 30, determine the map’s north orientation, follow the main avenue northward to its endpoint, and then use the legend to translate the final stand number into its corresponding name.}
    \label{fig:maps_example}
\end{figure}

\begin{figure}
    \centering
    \includegraphics[width=\linewidth]{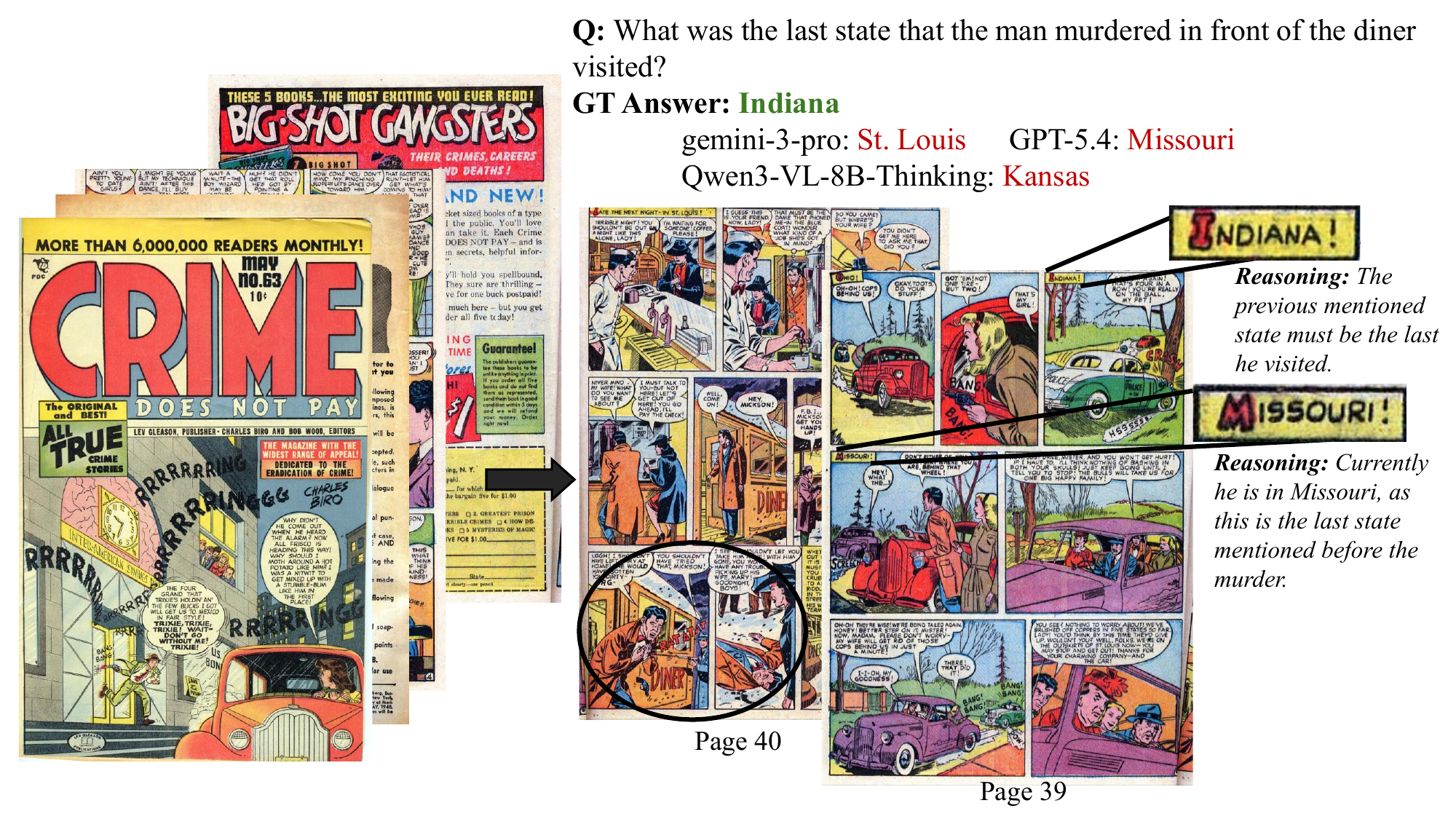}  
    \caption{Sample question from the comics domain: to answer it, the model must identify the page where the murder occurs, track the sequence of locations mentioned in the narrative panels leading up to that event, and determine the last state referenced before the state in which the crime is committed.}
    \label{fig:comics_example}
\end{figure}
DocVQA2026 is designed as a benchmark for multimodal reasoning over documents. Each question is associated with a single correct answer, although in some cases a small set of equivalent answer variants is accepted (e.g., \textit{green} and \textit{lime green}). The dataset is divided into two subsets, \textit{validation} and \textit{test}, and covers documents from eight domains:

\begin{itemize}
    \item \textbf{Business Reports:} Long documents containing tables, charts, and dense textual sections. Questions typically require navigating large contexts and extracting as well as synthesizing information from structured content.
    \item \textbf{Maps:} Historical maps from the 1500s to the 1800s, as well as modern road maps with high-resolution imagery. Questions focus on locating elements, following directions, and reasoning about spatial relationships, as illustrated in Figure~\ref{fig:maps_example}.
    \item \textbf{Engineering Drawings:} Detailed blueprints and technical schemes, many from NASA’s Apollo mission. Questions often involve identifying components, counting elements, and interpreting diagram structure.
    \item \textbf{Comics:} Colour comic books from the Golden Age of American comics. Questions target plot understanding, character tracking, and panel-level evidence, as shown in Figure~\ref{fig:comics_example}.
    \item \textbf{Slides:} Presentation documents on a variety of topics. Questions require interpreting visual elements, navigating across slides, and extracting relevant information.
    \item \textbf{Infographics:} Highly visual documents containing dense graphical content. Questions focus on charts, icons, and other visual elements distributed across the page.
    \item \textbf{Science Papers:} Multi-page academic papers. Questions target long-context reasoning and the identification of specific textual, tabular, or visual evidence.
    \item \textbf{Science Posters:} Scientific posters containing plots, diagrams, and experimental summaries. Questions are often layout-dependent and require combining visual and textual cues.
\end{itemize}

Questions are uniformly distributed across domains in both subsets. The validation set contains 10 questions per domain, for a total of 80 questions, while the test set contains 20 questions per domain, for a total of 160 questions. All questions were carefully created and validated by a team of human annotators.

\begin{figure}
    \centering
    \includegraphics[width=0.8\linewidth]{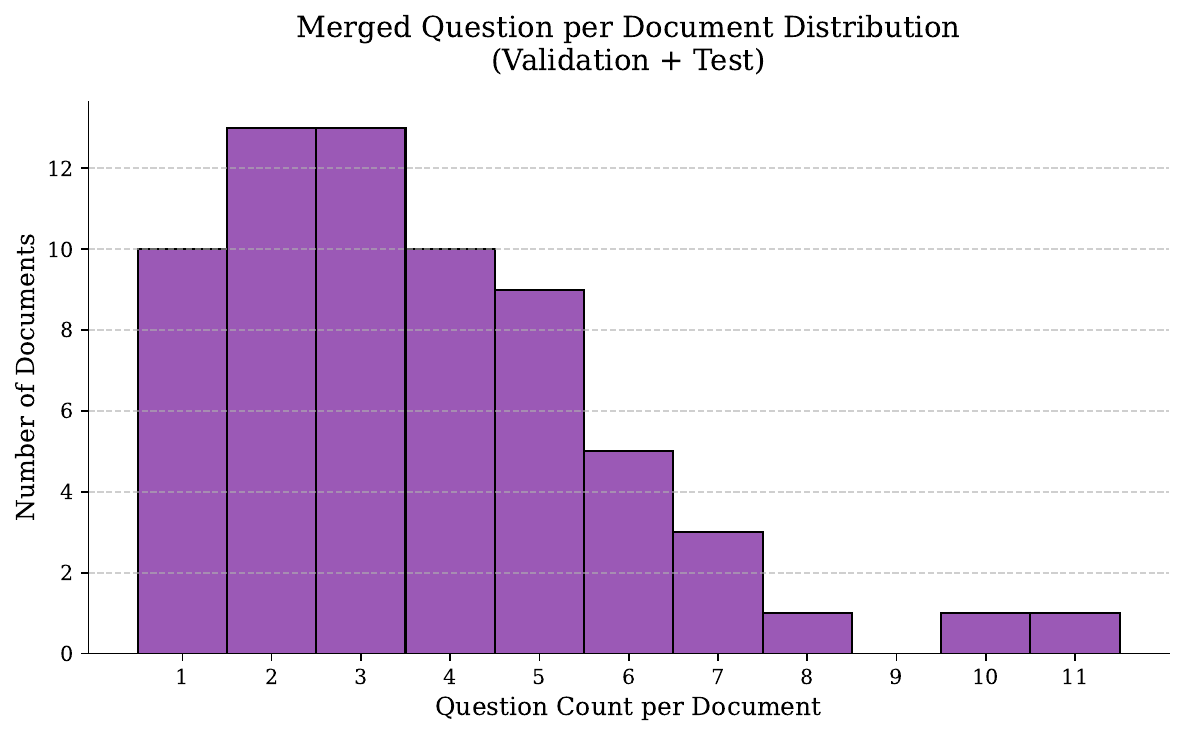}  
    \caption{Questions per Document Distribution (Validation+Test) }
    \label{fig:q_histo}
\end{figure}

The distribution of questions per document is more heterogeneous, as shown in Figure~\ref{fig:q_histo}. This ensures that each domain is evaluated over multiple documents rather than being dominated by only a few sources, leading to a more reliable benchmark.

A small fraction of the questions are intentionally unanswerable from the provided document and therefore have \texttt{Unknown} as the ground-truth answer. In the validation set, these account for $6.25\%$ of the questions, with the test set following a similar proportion. Such cases are included to evaluate whether systems can correctly abstain from answering when the required evidence is absent from the document. 

\subsection{Evaluation}

In order to facilitate evaluation, the DocVQA2026 dataset was designed to have exact, unambiguous answers that can be directly extracted from the documents. This design choice enables deterministic and reproducible evaluation without relying on LLM-as-a-judge paradigms. While LLM-based evaluation has recently gained popularity due to its scalability, it introduces several practical and methodological drawbacks. First, it imposes a non-negligible computational and financial burden on researchers wishing to evaluate their methods, as it typically requires access to large proprietary models and repeated inference over the full evaluation set. This creates an additional barrier to entry and hinders reproducibility, especially for researchers with limited resources.

More importantly, LLM-as-a-judge evaluations are known to suffer from calibration and bias issues. Recent work~\cite{lee2025correctly} shows that naive use of LLM judges leads to systematically biased estimates due to imperfect sensitivity and specificity, and that statistically sound evaluation requires additional calibration datasets and uncertainty-aware corrections.

\subsubsection{Standardized Answer Extraction}
To ensure reliable automatic assessment, the evaluation pipeline begins with a standardized prompting protocol that constrains the output format of the evaluated models. Models are instructed to apply strict formatting rules, such as normalizing units, converting dates to a standardized format, and isolating their final response prefixed by a final tag. After inference, the evaluation system parses the model output and extracts the textual span following this tag. If a model fails to follow the required output format and a valid final answer cannot be extracted, the prediction is counted as incorrect. This design choice reflects a realistic setting in which both answer correctness and instruction adherence are necessary for reliable extraction.

\subsubsection{Hierarchical Matching Strategy}
Once successfully extracted, each prediction is evaluated independently against its corresponding ground truth answer. For each question, the evaluation follows a hierarchical matching strategy:
\begin{itemize}
    \item \textbf{Strict Matching:} First, a strict matching criterion is applied to capture cases where precise equivalence is required. Predictions are considered correct if both the predicted answer and the ground truth can be interpreted as numerical quantities with associated units, and both the numeric value and the unit match exactly. Similarly, answers representing dates are considered correct if they correspond to the same calendar date after normalization. Version-like strings (e.g., 1.2.3) must also match exactly. This phase ensures that semantically critical formats are evaluated with high precision.
    \item \textbf{Relaxed Matching:} If none of the strict conditions are satisfied, the evaluation falls back to a relaxed textual matching scheme based on the Average Normalized Levenshtein Similarity (ANLS) \cite{biten2019scenetextvisualquestion,mathew2021docvqa}. Before comparison, both prediction and ground truth strings are normalized by converting text to lowercase, removing punctuation, and eliminating common articles. The similarity between a prediction $p$ and a ground truth answer $g$ is defined as:
    \begin{equation}
        \text{ANLS}(p, g) = 1 - \frac{d(p, g)}{\max(|p|, |g|)}
    \end{equation}
    where $d(p, g)$ denotes the Levenshtein edit distance, and $|p|, |g|$ represent the lengths of the respective strings. When multiple ground truth variants are available, the maximum similarity score across all candidates is considered. A prediction is deemed correct if this maximum similarity exceeds a threshold of 0.9.
\end{itemize}

\subsubsection{Evaluation Metrics}
Each sample is assigned a binary score based on correctness, with a value of 1 for correct predictions and 0 otherwise. The overall performance is measured using accuracy, defined as:
\begin{equation}
    \text{Accuracy} = \frac{1}{N} \sum_{i=1}^{N} \text{score}_{i}
\end{equation}
where $N$ is the total number of evaluated samples.

In addition to the global metric, performance is also analysed at the domain level. Each question is associated to a document from one of the eight domains, and domain-wise accuracy is computed as:
\begin{equation}
    \text{Accuracy}_{d} = \frac{\text{correct}_{d}}{\text{total}_{d}}
\end{equation}
where $\text{correct}_{d}$ and $\text{total}_{d}$ denote the number of correctly answered questions and the total number of questions within domain $d$, respectively. This evaluation protocol combines strict equivalence checks with robust similarity-based matching, ensuring both precision in structured answers and flexibility in natural language responses.

\subsection{Baselines}

We compare against a set of zero-shot Vision--Language Models (VLMs) baselines including both proprietary API-based systems and open-weight models executed locally. The proprietary baselines comprise \texttt{Gemini-3.1-Pro-preview}, \texttt{Gemini-3-Flash-preview}, \texttt{GPT5.2}, \texttt{GPT5.4}, and \texttt{GPT5.2-mini}~\cite{gemini31pro2026,gemini3flash2026,gpt54,gpt5mini}. All proprietary models were accessed through their corresponding APIs in March 2026 using a generation temperature of $1.0$, with high reasoning or thinking mode enabled when available. In parallel, we evaluated three models from the open-weight Qwen3-VL family~\cite{qwen3vl}, at sizes 2B, 8B, and 32B parameters . These models were run locally using \texttt{vLLM} on a server equipped with two NVIDIA RTX PRO 6000 Blackwell GPUs.

Many documents in the dataset span multiple pages with high-resolution images, causing their total visual token count to exceed the 256k maximum sequence length of the Qwen3-VL and Qwen3.5 series. To handle this, we compute the number of visual tokens each page would produce using the model's own tokenisation formula. When the total across all pages of a document exceeds our budget of 250k tokens, we distribute that budget evenly across pages and scale down each image according to the available budget per page.

\begin{figure}[h!]
\centering
\begin{minipage}{0.95\linewidth}
\footnotesize
\begin{tcolorbox}
\begin{Verbatim}[breaklines=true, breakanywhere=true]
ACT AS an expert Document Visual Question Answering (DocVQA) system. ANALYZE the provided images to extract precise information.

### MANDATORY RESPONSE RULES:
1. SOURCE ADHERENCE: If the question is unanswerable from the document, respond ONLY with "Unknown".
2. LIST FORMATTING: List multiple answers in order of appearance, separated by a comma and a single space (e.g., "Answer A, Answer B"). Do NOT use "and".
3. NUMBERS & UNITS:
   - Convert units to their standardized abbreviation (e.g., use "kg" not "kilograms", "m" not "meters").
   - Place a single space between the number and the unit (e.g., "50 kg", "10 USD").
4. PERCENTAGES: For percentages, attach the '%' symbol directly to the number with NO space (e.g., "50%", not "50 %").
5. DATE FORMATTING: Convert all dates to YYYY-MM-DD format (e.g., convert "Jan 1st 24" to "2024-01-01").
6. DECIMAL FORMATTING: Decimals should be separated by a single period (e.g., "3.14", not "3,14").
7. THOUSANDS SEPARATOR: Do NOT use commas as thousands separators (e.g., "1000", not "1,000").
8. NO FILLER: Output ONLY the result. Do not frame with sentences like "The answer is...".

### REASONING PROTOCOL:
1. Perform exhaustive step-by-step reasoning to locate and verify the data.
2. Verify if the data contains a date, number, or unit.
3. Step-by-step, transform the data to match the MANDATORY RESPONSE RULES.

### OUTPUT FORMAT:
FINAL ANSWER: [Your exact formatted answer]
\end{Verbatim}
\end{tcolorbox}
\end{minipage}
\caption{Prompt used for all baseline models.}
\label{fig:prompt}
\end{figure}

To ensure a fair comparison, the same instruction prompt was used for all baselines, available in Figure~\ref{fig:prompt}. The prompt frames the model as an expert Document Visual Question Answering system and explicitly requires the answer to be derived only from the provided document images. It also imposes strict formatting constraints intended to reduce ambiguity during evaluation. In particular, the prompt instructs models to return \texttt{Unknown} when the answer cannot be inferred from the document, to format multi-item answers as comma-separated lists in reading order, to normalize units using standard abbreviations, to format percentages without intervening spaces, to convert dates to the \texttt{YYYY-MM-DD} format, to use period-based decimal notation, to avoid thousands separators, and to omit any explanatory or conversational text. The prompt additionally encourages step-by-step internal reasoning and requires the final response to be provided in a canonical form prefixed by \texttt{FINAL ANSWER:}. This standardized prompting protocol is intended to minimize performance differences caused by output formatting rather than actual document understanding.

For evaluation, only the final extracted answer is retained. More precisely, after inference, the system parses the model output and extracts the content following the \texttt{FINAL ANSWER:} tag. This extracted span is then passed to the evaluation pipeline. If a model fails to follow the required output format and a valid final answer cannot be extracted, the prediction is counted as incorrect. This design choice reflects a realistic setting in which both answer correctness and instruction adherence are necessary for reliable automatic assessment.

\section{Competition Protocol}
The competition was conducted through the Robust Reading Competition (RRC) portal\footnote{\url{https://rrc.cvc.uab.es/?ch=34}}, where participants were required to register in order to take part in the evaluation process. The DocVQA2026 dataset was made publicly available via Hugging Face with ground-truth annotations provided only for the validation set, while the test set included questions without the answers. The validation set was released at the end of January, followed by the release of the test set at the beginning of March. The competition concluded on April 3rd, giving participants a three-week window to evaluate their methods on the test data.

To support system development, we provided evaluation code that could be used locally on the validation set, and in addition, an official evaluation server for validation was accessible through the RRC platform. The test set could only be evaluated through this server, ensuring a controlled and fair benchmarking procedure. The main evaluation metric is overall accuracy computed across all domains, with results also reported separately for each domain to provide a more detailed analysis of system performance.

Participants were allowed to submit multiple entries; however, each submission was expected to represent a substantially different approach, such as the use of different models or significant methodological changes. Submissions were monitored to prevent misuse of the evaluation system, particularly attempts to overfit or optimize directly on the test set.

All methods are presented within a single leaderboard, but participants were required at submission time to specify the category of their approach based on the total number of parameters. This count includes all parameters involved in the system, regardless of whether they are active at inference time, and also accounts for all models used in agentic pipelines. In cases where private or API-based models with undisclosed parameter counts were used, submissions were assigned to the largest category by default. Additional details about each approach were collected by requesting a short report from participants.

In total, 20 submissions where accepted as valid, coming from 8 different teams.

\section{Submitted Methods and Results}

\begin{longtable}{>{\raggedright\arraybackslash}p{1.1cm}
                  >{\raggedright\arraybackslash}p{5.0cm}
                  >{\centering\arraybackslash}p{1.5cm}
                  >{\raggedright\arraybackslash}p{2cm}
                  >{\centering\arraybackslash}p{2cm}}
\caption{Overall DocVQA competition results across all submitted methods, grouped in a single table spanning the three parameter bands (\(\leq\)8B, 8--35B, and \(>\)35B). Accuracy is shown on a 0--100 percentage scale. Boldface indicates the winner in each parameter band, with a tie in the 8--35B band. Light-gray rows correspond to official baselines.}
\label{tab:docvqa_all_methods_total}\\
\toprule
Cat. & Method & Params & Type & Accuracy\\
\midrule
\endfirsthead

\toprule
Cat. & Method & Params & Type & Accuracy\\
\midrule
\endhead

\midrule
\multicolumn{5}{r}{\emph{Continued on next page}}\\
\endfoot

\bottomrule
\endlastfoot

\CatLarge & \textbf{Luna + IDA + Multi-Agent Ensemble + with FT} &  & Multi-agent ensemble & \textbf{60.00} \\
\CatLarge & Luna + IDA + Multi-Agent Ensemble &  & Multi-agent ensemble & 55.63 \\
\CatLarge & Uni-Parser Tools + Gemini-3.1-Pro &  & Parser/OCR\ zero-shot & 46.25 \\
\CatLarge & ARGUS\_Gemini\_3.1\_Pro &  & Agentic RAG & 43.75 \\
\rowcolor{baselinegray} \CatLarge & Gemini-3.1-Pro-preview &  & Single-model zero-shot & 37.50 \\
\rowcolor{baselinegray} \CatLarge & Gemini-3-Flash-preview &  & Single-model zero-shot & 35.63 \\
\CatMid   & \textbf{ARGUS\_Qwen3.5\_27B} & \textbf{27B} & Agentic RAG & \textbf{35.63} \\
\CatMid   & \textbf{Perceive-Reason-Code} & \textbf{27B} & Agentic RAG & \textbf{35.63} \\
\CatMid   & Uni-Parser Tools + Qwen3.5-27B & 27B & Parser/OCR\ zero-shot & 31.88 \\
\CatLarge & rocky mode &  & Agentic RAG & 27.50 \\
\rowcolor{baselinegray}\CatLarge & GPT-5.2 &  & Single-model zero-shot & 26.88 \\
\rowcolor{baselinegray}\CatLarge & GPT-5.4 &  & Single-model zero-shot & 25.63 \\
\CatMid   & qwen35vl\_27b\_sft & 27B & VLM FT & 23.75 \\
\CatMid   & HALLEY & 9B & Parser/OCR\ zero-shot & 20.00 \\
\CatSmall & \textbf{Uni-Parser Tools + Qwen3.5-4B} & \textbf{4B} & Parser/OCR\ zero-shot & \textbf{18.75} \\
\rowcolor{baselinegray} \CatLarge & GPT-5-mini &  & Single-model zero-shot & 17.50 \\
\CatMid   & qwen3-vl-32b & 32B & Single-model zero-shot & 15.63 \\
\rowcolor{baselinegray} \CatMid   & Qwen3-VL-32B-Thinking & 32B & Single-model zero-shot & 14.38 \\
\CatSmall & ARGUS\_Qwen3-VL-8B-Thinking & 8B & Agentic RAG & 14.38 \\
\CatSmall & qwen3vl-8b-sft & 8B & VLM FT & 10.63 \\
\rowcolor{baselinegray} \CatSmall & Qwen3-VL-8B-Thinking & 8B & Single-model zero-shot & 9.38 \\
\CatSmall & Pushing 8B model & 8B & Agentic RAG & 9.38 \\
\CatLarge & Multi-Agent RAG for Document VQA &  & Agentic RAG & 6.25 \\
\CatSmall & Zero-shot approach with tiny models & 8B & Single-model zero-shot & 5.00 \\
\rowcolor{baselinegray} \CatSmall & Qwen3-VL-2B-Thinking & 2B & Single-model zero-shot & 1.88 \\
\CatSmall & Vision Encoder OCR Model &  & Single-model zero-shot & 1.25 \\

\end{longtable}

Table~\ref{tab:docvqa_all_methods_total} summarizes the overall test-set results of all submitted methods. Although winners were selected independently within the three parameter bands (\(\leq\)8B, 8--35B, and \(>\)35B), all systems are shown together to facilitate comparison across scales. This joint view is particularly informative in this benchmark, where several carefully engineered smaller systems outperform substantially larger zero-shot baselines. The table therefore highlights not only the strongest methods within each category, but also the extent to which system design, inference-time orchestration, and document-specific preprocessing can compensate for model scale.

To make the leaderboard easier to interpret, we group submissions into a small number of broad method types derived from the system reports. These labels are descriptive rather than absolute, since many submissions combine several ideas. Broadly, the submitted systems fall into five families: \emph{single-model zero-shot} methods, \emph{parser/OCR-augmented zero-shot} methods, \emph{agentic RAG} methods, \emph{multi-agent ensembles}, and \emph{fine-tuned VLMs}. A clear pattern across the leaderboard is that the strongest entries move beyond single-pass end-to-end prompting and instead add structure through better evidence extraction, page selection, verification, or aggregation of complementary model perspectives.

\subsection{Method Landscape}

The submitted systems can be grouped into five broad method families:

\begin{itemize}
    \item \textbf{Single-model zero-shot:} a single VLM answers directly from the document images, with no task-specific adaptation. This family includes the official baselines and provides a useful reference point for out-of-the-box document reasoning.

    \item \textbf{Parser/OCR-augmented zero-shot:} zero-shot inference is supported with OCR, document parsing, or adaptive cropping. These methods aim to improve evidence access, especially in text-heavy or dense layouts.

    \item \textbf{Agentic RAG:} the system actively searches or navigates the document before answering, for example through retrieval, cropping, tool use, or iterative exploration.

    \item \textbf{Multi-agent ensembles:} several readers or reasoning stages are combined, often with voting or conflict resolution, to aggregate complementary strengths.

    \item \textbf{Fine-tuned VLMs:} the underlying multimodal model is adapted with task-specific supervision rather than relying only on inference-time engineering.
\end{itemize}

Overall, the strongest submissions move beyond single-pass prompting and instead structure the interaction between document evidence and model reasoning through retrieval, verification, or aggregation.

\subsection{Representative Systems}

\subsubsection{Luna + IDA + Multi-Agent Ensemble}

The winning submission in the \(>\)35B category, \emph{Luna + IDA + Multi-Agent Ensemble + with FT}, is the clearest example of large-scale orchestration. It is a pipeline combining IDA, a layout-aware parser, multiple VLM reader agents, and a final reasoning agent that resolves conflicts across the extracted evidence. The system also reformulates questions for different readers and applies domain-dependent trust priorities, favoring OCR-grounded evidence for text-heavy documents and visually stronger readers for image-heavy ones. More than any other submission, Luna illustrates the effectiveness of decomposing DocVQA into specialized stages rather than relying on a single end-to-end pass.

\subsubsection{ARGUS and Perceive-Reason-Code}

The 8--35B category ends in a tie between \emph{ARGUS\_Qwen3.5\_27B} and \emph{Perceive-Reason-Code}, both reaching \(35.63\) accuracy, but through very different designs. ARGUS is centered on OCR, hybrid retrieval, and adaptive context selection: it combines lexical retrieval over OCR text with visual retrieval over page images, then selects either the full document or a reduced evidence set depending on the case. For some documents, it also uses an intermediate investigation stage before final answering.

By contrast, \emph{Perceive-Reason-Code} frames DocVQA as a code-generation agent problem. A single Qwen3.5-27B model operates inside a Python REPL, using OCR-based search and visual inspection tools to explore the document, inspect candidate regions, and compute the final answer. The tie between these two methods is one of the most interesting results in the leaderboard, as it shows that similar overall performance can be reached through either retrieval-and-context orchestration or a more explicitly agentic code-execution framework.

\subsubsection{HALLEY}

This method deserves special mention as an efficiency-oriented system. Although submitted in the 8--35B category, it is based on Qwen3.5-9B and still reaches \(20.00\) accuracy, outperforming the 32B zero-shot Qwen variants and the GPT-5-mini baseline. Its report attributes this to a four-level vision cascade combining whole-page inference, OCR injection, adaptive tiling, and tiled OCR prompting, together with category-aware layouts and custom multi-page question strategies. HALLEY is therefore a strong example of a smaller model remaining competitive through careful system engineering.

\subsubsection{Uni-Parser Tools}

This family of methods is representative of parser/OCR-augmented zero-shot systems. It is based on a pipeline based on adaptive-resolution cropping, prompt engineering, self-judgment, retries, and in some variants mixed-model voting. These methods perform strongly across scales, including the winning submission in the \(\leq\)8B category and a competitive \(>\)35B entry with Gemini-3.1-Pro. This family illustrates a broader trend in the competition: even without fine-tuning, structured preprocessing and verification can substantially improve performance over direct prompting.

\subsection{Performance Across Document Domains}

\begin{figure}[t]
    \centering
    \includegraphics[width=\linewidth]{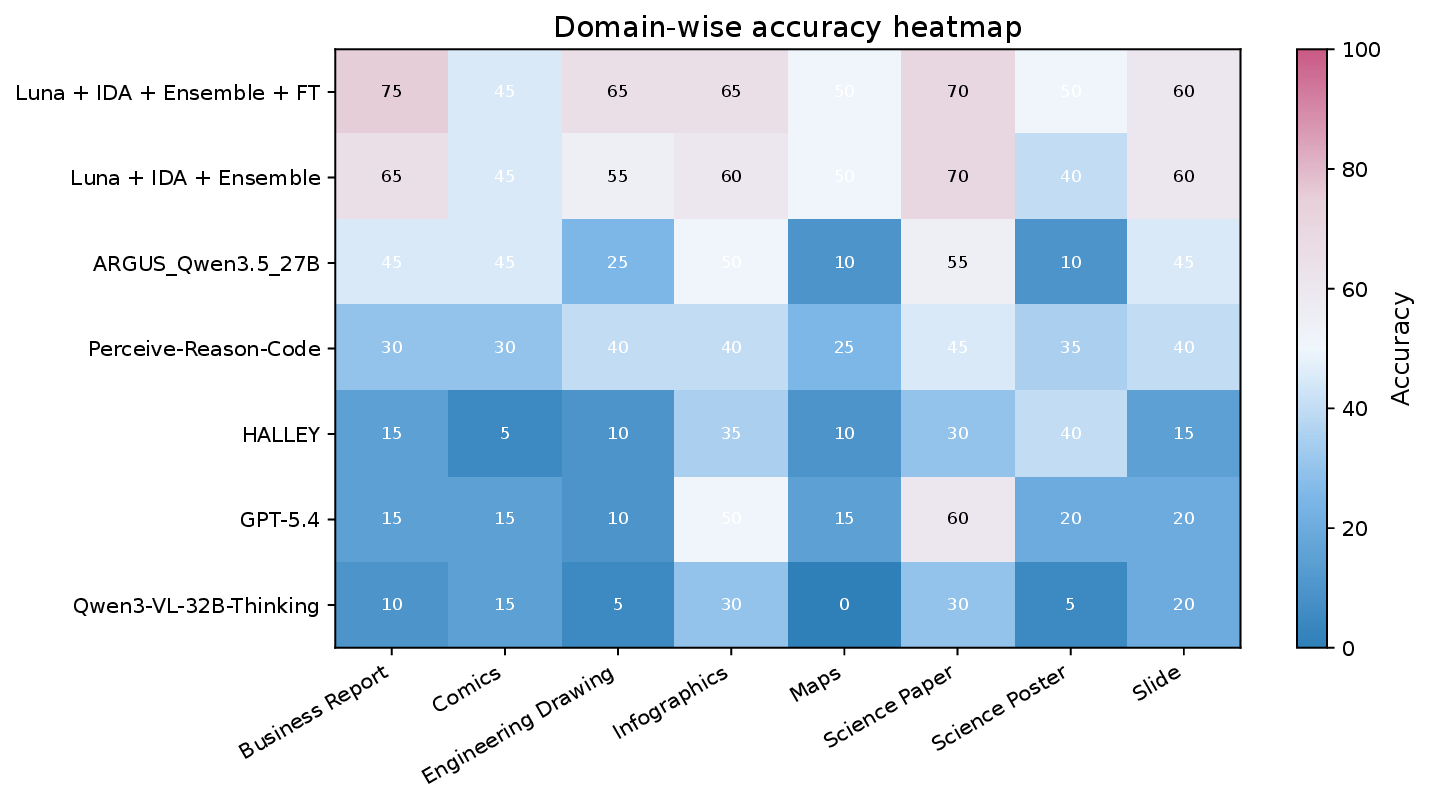}
    \caption{Domain-wise accuracy for selected representative methods. The figure highlights differences in difficulty across domains and shows that the strongest systems are more balanced than the zero-shot baselines across categories.}
    \label{fig:domain_heatmap}
\end{figure}

To better understand the strongest submissions, Figure~\ref{fig:domain_heatmap} compares domain-wise accuracy for a selected subset of representative methods: the two top \(>\)35B submissions, the two tied 27B methods, HALLEY as a compact but competitive 9B system, and two zero-shot baselines. This compact view reveals both domain difficulty and differences in method behavior.

A first observation is that performance varies substantially by domain. Business reports, science papers, and slides are among the best-performing categories for the top systems, while maps remain difficult across nearly all methods. Engineering drawings and comics are also challenging for many submissions, especially for zero-shot baselines. These results confirm that DocVQA2026 challenges multiple abilities at once, from structured evidence extraction to spatial and narrative reasoning.

The comparison also reinforces the importance of system design. The two Luna variants are the strongest and most balanced methods across domains, suggesting that their multi-agent decomposition improves both average performance and robustness. In contrast, the zero-shot baselines are much less consistent: GPT-5.4 is competitive on some domains, but weak on others, while Qwen3-VL-32B-Thinking remains generally limited despite its size. HALLEY provides a useful counterpoint, as its 9B system remains competitive on several domains and outperforms larger zero-shot baselines in some cases, showing that careful inference-time engineering can partly compensate for reduced scale.

\begin{figure}[t]
    \centering
    \includegraphics[width=\linewidth]{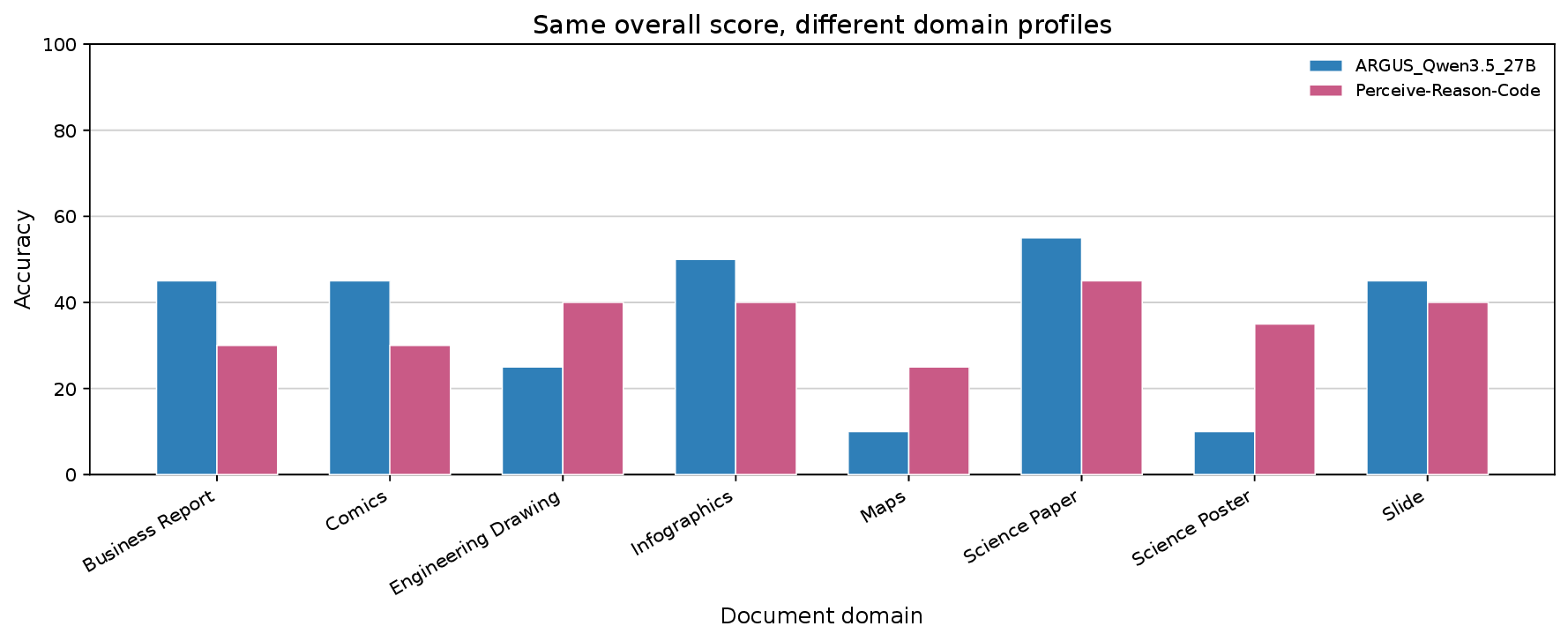}
    \caption{Domain-wise comparison of the two tied methods in the 8--35B category. Although \emph{ARGUS\_Qwen3.5\_27B} and \emph{Perceive-Reason-Code} achieve the same overall accuracy, they do so through different domain strengths.}
    \label{fig:tie_analysis}
\end{figure}

An especially informative comparison is the tie in the 8--35B category between \emph{ARGUS\_Qwen3.5\_27B} and \emph{Perceive-Reason-Code}. Although both methods obtain the same overall accuracy, their domain profiles differ noticeably, as shown in Figure~\ref{fig:tie_analysis}. ARGUS is stronger on business reports, comics, infographics, science papers, and slides, while Perceive-Reason-Code performs better on engineering drawings, maps, and science posters.

This tie shows that overall accuracy alone does not fully capture system behaviour on DocVQA2026. ARGUS appears stronger in text-heavy and mixed-layout settings, consistent with its emphasis on OCR, retrieval, and adaptive context selection. Perceive-Reason-Code, in contrast, gains ground in domains where interactive inspection may be more useful, such as engineering drawings or maps. Overall, the domain-wise analysis confirms three main conclusions: some document types remain substantially harder than others, structured systems are more robust than plain zero-shot baselines, and methods with similar overall scores may still have very different strengths.

\section{Conclusions}

This report presents the results of the ICDAR2026 Competition on Multimodal Reasoning over Documents in Multiple Domains. The submitted methods show that this benchmark remains far from saturated: while zero-shot VLMs provide a strong starting point, the best-performing systems consistently rely on additional structure, such as OCR and parsing, retrieval, adaptive evidence selection, self-verification, or multi-agent orchestration.

The results also highlight several open challenges for document understanding. Maps, engineering drawings, and comics remain particularly difficult, suggesting that robust multimodal reasoning over specialized layouts, small visual details, and narrative or spatial dependencies is still an open problem. More broadly, the competition shows that overall accuracy alone does not fully characterize system behaviour, as methods with similar aggregate scores can exhibit very different strengths across domains. This reinforces the importance of diverse benchmarks and domain-wise evaluation for measuring progress in document understanding.

Beyond the competition itself, we expect DocVQA2026 to remain a useful resource for the community. The dataset, evaluation code, and official server will continue to be available after the end of the challenge, allowing new methods to be evaluated under the same protocol. In fact, since the competition closed, many additional methods have already been tested on the benchmark, further confirming the continued interest of the community in this task.

\section*{Acknowledgements}
This publication has been supported by the Consolidated Research Group 2021 SGR 01559 from the Research and University Department of the Catalan Government, and by project PID2023-146426NB-100 funded by MCIU\allowbreak/AEI\allowbreak/\allowbreak10.13039\allowbreak/\allowbreak501100011033 and FEDER, UE.\newline

This work has been funded by the European Large Open Multi-Modal Foundation Models For Robust Generalization On Arbitrary Data Streams (ELLIOT) from the European Union’s Horizon Europe programme under grant agreements No 101070617 and 101214398. With the support of the FI-STEP predoctoral grant program from the Department of Research and Universities of the Generalitat de Catalunya and co-financing by the European Social Fund Plus (2025STEP00060).\newline 

%
\bibliographystyle{splncs04}
\bibliography{mybibliography}
\end{document}